\documentclass[twoside]{article}
\usepackage{epsfig,latexsym,amsbsy,amssymb,amsmath,color, url,booktabs, multirow}
\usepackage{graphicx,xspace}
\usepackage{longtable}
\usepackage{mathtools}
\usepackage{algorithm,algpseudocode}
\usepackage{enumitem}  
\usepackage[round]{natbib}
\usepackage{fullpage} 

\newcommand{\commentOut}[1]{} 

\newcommand{\mat}[1]{\mathbf{#1}}
\renewcommand{\vec}[1]{ \mathbf{#1} } 
\newcommand{\vecS}[1]{\boldsymbol{ #1 }  } 

\newcommand{\matentry}[3]{{\mathrm #1_{#2,#3}}}
\newcommand{\matcol}[2]{\mat{#1}_{\cdot,#2}}

\newcommand{\K}{\mat{K}}

\newcommand{\X}{\mat{X}}

\newcommand{\calD}{\mathcal{D}} 
\newcommand{\calL}{\mathcal{L}}

\newcommand{\calO}{\mathcal{O}}

\providecommand{\norm}[1]{\lVert#1\rVert}

\newcommand{\expectation}[2]{ \mathbb{E}_{#1}{\left[#2\right]} }

\newcommand{\Normal}{\mathcal{N}}

\newcommand{\gp}{\mathcal{GP}}
\newcommand{\kernel}{\kappa}
\newcommand{\meanfunc}{\mu}

\newcommand{\deriv}[2]{\frac{\partial{#1}}{\partial{#2}}}

\newcommand{\gradient}{\nabla}

\newcommand{\defeq}{\stackrel{\text{\tiny def}}{=}}

\newcommand{\mth}{\mathrm{th}}

\newcommand{\bigO}{\calO}

\newcommand{\kl}[2]{\mathrm{KL}(#1 \lVert #2)}

\newcommand{\loocv}{\name{loo-cv}}

\newcommand{\name}[1]{{\textsc{#1}}\xspace}

\newcommand{\margmean}[1]{b_{kn{#1}}}
\newcommand{\margvar}[1]{\sigma^2_{kn{#1}}}
\newcommand{\margstd}[1]{\sigma_{kn{#1}}}
\newcommand{\batchset}{\Omega}
\newcommand{\batchsize}{B}
\newcommand{\degree}{d}
\newcommand{\depth}{l}
 \newcommand{\arckernel}[4]{\kernel^{(#1)}_{#2}(#3, #4)}
 \newcommand{\arcangle}{\phi}
 
\newcommand{\ynotn}{\y_{\neg n}}
\newcommand{\Dnotn}{\calD_{\neg n}}
\newcommand{\Xnotn}{\X_{\neg n}}

\definecolor{mygreen}{rgb}{0.2, 0.7, 0.2}
\definecolor{myorange}{rgb}{0.9, 0.5, 0.0}

\newcommand{\gpu}{\name{gpu}}

\newcommand{\cifar}{\name{cifar10}}

\newcommand{\rectangle}{\name{rectangles-image}}
\newcommand{\mnisteight}{\textsc{mnist}8\textsc{m}\xspace}

\newcommand{\sarcos}{\name{sarcos}} 
\newcommand{\mnist}{\name{mnist}} 

\newcommand{\rmsprop}{\textsc{rmsp}rop\xspace}
\newcommand{\autogp}{\textsc{a}uto\textsc{gp}\xspace}
\newcommand{\arccosine}{\name{arc-cosine}}
\newcommand{\dbnthree}{\name{DBN}}
\newcommand{\rbf}{\name{rbf}}
\newcommand{\svm}{\name{svm}}
\newcommand{\dl}{\name{dl}}

\newcommand{\mcmc}{\name{mcmc}}
\newcommand{\gptext}{{\sc gp}\xspace}
\newcommand{\savigp}{{\sc savigp}\xspace}

\newcommand{\ard}{\name{ard}}
\newcommand{\rbfard}{\name{rbf-ard}}

\newcommand{\gprn}{\name{gprn}}

\newcommand{\mc}{\name{mc}}

\newcommand{\sgd}{\name{sgd}}

\newcommand{\looobj}{\calL_{\text{oo}}}
\newcommand{\looobjhat}{\widehat{\calL}_{\text{oo}}}

\newcommand{\elbohat}{\widehat{\calL}_{\text{elbo}}}
\newcommand{\elbo}{\calL_{\text{elbo}}}

\newcommand{\elltermhat}{\widehat{\calL}_{\text{ell}}}  

\newcommand{\elltermhatn}{\elltermhat^{(n)}}

\newcommand{\msse}{\name{msse}}

\newcommand{\nlp}{\name{nlp}}

\newcommand{\er}{\name{er}}

\newcommand{\param}{\vecS{\eta}}
\newcommand{\hyperparam}{\vecS{\theta}}
\newcommand{\hyper}{\hyperparam}

\newcommand{\varparam}{\llambda}
\newcommand{\pcond}[2]{p(#1  | #2 )}

\newcommand{\bs}{\boldsymbol}
\newcommand{\llambda}{\bs{\lambda}}

\newcommand{\mydot}{\cdot}

\newcommand{\s}{S} 
\newcommand{\n}{N} 
\newcommand{\m}{M} 
\newcommand{\p}{P} 
\newcommand{\q}{Q} 
\renewcommand{\d}{D} 
\renewcommand{\k}{K} 
\newcommand{\y}{\vec{y}} 

\newcommand{\yn}{\y_{n}} 

\newcommand{\f}{\vec{f}} 
\newcommand{\fn}{\f_{n \mydot}} 
\newcommand{\fj}{\f_{\mydot j}} 
\renewcommand{\u}{\vec{u}} 
\newcommand{\uj}{\vec{u}_{\mydot j} } 
\newcommand{\x}{\vec{x}}
\newcommand{\z}{\vec{z}}
\renewcommand{\X}{\mat{X}}
\newcommand{\Zj}{\mat{Z}_j}
\newcommand{\xstar}{\x_{\star}}

\newcommand{\ystar}{\y_{\star}}

\renewcommand{\K}{\mat{K}}
\newcommand{\Gram}[2]{\K_{#1#2}^j}
\newcommand{\Kzz}{\K_{\z\z}^j} 
\newcommand{\Kxx}{\K_{\x\x}^j}
\newcommand{\Kxz}{\K_{\x\z}^j} 
\newcommand{\Kzx}{\K_{\z\x}^j} 

\newcommand{\Kzzinv}{(\K_{\z\z}^j)^{-1}} 
\newcommand{\Aj}{\mat{A}_j}
\newcommand{\priormean}{\tilde{\vecS{\mu}}_j} 
\newcommand{\priorcov}{\widetilde{\K}_j}
\newcommand{\postmean}[1]{\vec{m}_{#1}}
\newcommand{\postcov}[1]{\mat{S}_{#1}}

\newcommand{\ajn}{\vec{a}_{jn}}

\newcommand{\qjoint}{q(\f, \u | \llambda)}
\newcommand{\qu}{q(\u | \llambda)}
  
\newcommand{\qf}{q(\f | \llambda)}

\newcommand{\qknf}{q_{k(n)}(\fn | \llambda_k )}

\newcommand{\qfmean}[1]{\vec{b}_{#1}}        
\newcommand{\qfcov}[1]{\mat{\Sigma}_{#1}}        

\renewcommand{\matentry}[3]{{[#1]_{#2,#3}}}

\renewcommand{\matcol}[2]{[{#1}]_{:,#2}}

\newcommand{\grad}{\gradient}

\newcommand{\iid}{i.i.d.\xspace~}

\newcommand{\footremember}[2]{%
	\footnote{#2}
	\newcounter{#1}
	\setcounter{#1}{\value{footnote}}%
}
\newcommand{\footrecall}[1]{%
	\footnotemark[\value{#1}]%
}

\title{AutoGP: Exploring the Capabilities and Limitations of\\ Gaussian Process Models}

\author{Karl Krauth\footremember{unsw} {School of Computer Science and Engineering,
		The University of New South Wales, Sydney NSW 2052.}
	\and  
	Edwin V.~Bonilla\footrecall{unsw}
	\and  
	Kurt Cutajar\footremember{eurecom}{Department of Data Science, EURECOM, France.}
	\and
	Maurizio Filippone\footrecall{eurecom}
} 

\begin{document}

%

%

%
%
%

\maketitle

\begin{abstract}
We investigate the capabilities and limitations of Gaussian process models by jointly exploring three complementary directions: 
(i) scalable and statistically efficient inference; 
(ii) flexible kernels;
and 
(iii) objective functions for hyperparameter learning alternative to the marginal likelihood. 
Our approach outperforms all previously reported \gptext methods on the standard  \mnist dataset; 
performs comparatively to previous kernel-based methods using the \rectangle dataset;
and 
breaks the 1\% error-rate barrier in \gptext models using the \mnisteight dataset,   
showing along the way the scalability of our method at unprecedented scale 
for \gptext models (8 million observations) in classification problems.	
 Overall, our approach represents a significant breakthrough in kernel methods and 
 \gptext models, bridging the gap between deep learning approaches and 
 kernel machines.

\end{abstract}

\section{INTRODUCTION}

Recent advances in deep learning \citep[\dl;][]{lecun-et-al-nature-2015} have revolutionized the application of  machine learning in areas such as 
computer vision \citep{krizhevsky-et-al-nips-2012},  
speech recognition \citep{hinton2012deep} and natural language processing \citep{collobert2008unified}. 
Although certain kernel-based methods have also been successful in such domains \citep{chonips2009,mairal2014convolutional}, 
it is still unclear whether these methods 
can indeed catch up with the recent \dl breakthroughs. 
 
Aside from the benefits obtained from using compositional representations, we believe that the 
main components contributing to the success of \dl techniques are:  
(i)	their scalability to large datasets and efficient computation via \gpu{s};
(ii) their large representational power; and
(iii) the use of well-targeted objective functions for the problem at hand. 

In the kernel world, Gaussian process \citep[\gptext;][]{rasmussen-williams-book} models  
are attractive because they are elegant Bayesian  nonparametric 
approaches to learning from data. 
Nevertheless, besides the limitations intrinsic to local kernel machines \citep{bengio2005curse}, 
it is clear that \gptext-based methods have not fully explored the desirable criteria 
highlighted above. 

Firstly, with regards to (i) scalability, despite recent advances in 
inducing-variable approaches and variational inference in \gptext models 
\citep{titsias2009variational,hensmangaussian,hensman-et-al-aistats-2015,dezfouli-bonilla-nips-2015}, 
the study of truly large datasets in problems other than regression and the 
investigation of \gpu-based acceleration in \gptext models are still 
under-explored areas. We note that these  issues are also shared by 
non-probabilistic kernel methods such as support vector machines
\citep[\svm{s};][]{scholkopf2001learning}.
 
Furthermore, concerning (ii) their representational power, 
kernel methods have been plagued by the overuse of very limited kernels such as 
the squared exponential kernel, also known as the radial-basis-function (\rbf) kernel.
Even worse, in some \gptext-based approaches, and more commonly in \svm{s}, these have been 
limited to having a single length-scale (a single bandwidth in \svm parlance) shared across all 
inputs instead of using different length-scales for each dimensions. The latter approach is commonly known
in the \gptext literature as automatic relevance determination 
\citep[\ard;][]{Mackay94}. 

Finally, and perhaps most importantly,  \gptext methods have largely ignored defining (iii) well-targeted objective functions, and instead focused heavily on optimizing 
 the marginal likelihood \cite[][Ch.~5]{rasmussen-williams-book}.
 This includes variational approaches 
 that optimize a lower bound of the marginal likelihood \citep{titsias2009variational}. 
Although different approaches 
 for hyperparameter learning in \gptext models are discussed by 
 \citet[][\S 5.4.2]{rasmussen-williams-book}, and indeed carried out by 
 \citet{sundararajan2001predictive,sundararajan07,sundararajan08} and more 
 recently by \citet{vehtari2016bayesian}, their performance results are somewhat limited by their disregard for other directions of improvement (scalability and greater representational power) mentioned above.

In this work we push the capabilities of \gptext-models, while also investigating their limitations,  
by addressing the above issues jointly.  In particular, 
\begin{enumerate}
	\item 
		we develop scalable and statistically efficient inference methods for \gptext models using 
		\gpu computation and stochastic variational inference along with the reparameterization trick \citep{kingma2013auto};  
	\item
		we investigate the flexibility of models using \ard kernels for high-dimensional 
		problems, as well as ``deep'' \arccosine kernels \citep{chonips2009}; and
	\item
		we study the impact of employing a leave-one-out-based objective function on hyperparameter learning.  
\end{enumerate}

As we rely on automatic differentiation \citep{baydin2015automatic} for the implementation of our model in TensorFlow \citep{tensorflow2015-whitepaper},
we refer to our approach as \autogp. While we thoroughly evaluate our claimed contributions on various 
regression and classification problems, our most significant results show that \autogp:
\begin{itemize}
	\item  has superior performance  to 
	all previously reported \gptext-based methods on the standard  \mnist dataset;
	\item
	achieves comparable performance to previous 
	kernel-based methods using the \rectangle dataset;  and 
	\item
	breaks the 1\% error-rate barrier in \gptext models using the \mnisteight dataset,   
	showing along the way the scalability of our approach at unprecedented scale 
	for \gptext models (8 million observations) in multiclass classification problems.
\end{itemize} 

Overall, \autogp represents a significant breakthrough in kernel methods in general and 
in \gptext models in particular, bridging the gap between deep learning approaches and 
kernel machines. 


\section{Related Work}

The majority of works related to scalable kernel machines primarily target issues pertaining to the storage and computational requirements of algebraic operations involving kernel matrices.
Low-rank approximations are ubiquitous among these approaches,  with  
the Nystr\"om approximation being one of the most popular methods in this category \citep{Williams00b}. 
Nevertheless,  most of these  approximations can be understood within the unifying probabilistic 
framework of \citet{quinonero2005unifying}.


The optimization of inducing inputs using Nystr\"om approximations for \gptext{s} made significant progress with the work on sparse-\gptext variational inference in \citet{titsias2009variational}.
This exposed the possibility to develop stochastic gradient optimization for \gptext models \citep{hensmangaussian}
and 
handle non-Gaussian likelihoods \citep{nguyen-bonilla-nips-2014,dezfouli-bonilla-nips-2015,hensman-et-al-aistats-2015,sheth2015sparse}, which has sparked interest
in  the implementation of scalable inference frameworks such as those in \citet{gpflow} and \citet{deisenroth-ng-icml-2015}.
These developments greatly improved the generality and applicability of \gptext models to a variety of applications, and our work follows from this line of ideas. 

Other independent works have focused on kernel design \citep{Wilson13, chonips2009,mairal2014convolutional,Wilson16}.  
This has sparked some debate as to whether kernel machines can actually compete with deep nets. 
Evidence suggests that this is possible; notable examples include the work by \citet{Huang14,Lu14}.
These works also provide insights into the aspects that make kernel methods less competitive to deep neural networks, namely lack of application specific kernels, and scalability at the price of poor approximations. 
These observations are corroborated by our experiments, which report how the combination of these factors impacts performance of \gptext{s}.

Because \gptext{s} are probabilistic kernel machines, it is natural to target the optimization of the marginal likelihood. 
However, alternative objective functions have also been considered, in particular the 
leave-one-out cross-validation (\loocv) error. 
Originally this was done by \citet{sundararajan2001predictive} for regression;    
 later extended by \citet{sundararajan07} to deal with  sparse \gptext formulations;  
 	and  broadened  by 
 	\citet{sundararajan08,vehtari2016bayesian} to handle non-Gaussian likelihoods.  
While the results in these papers seem inconclusive as to whether this can generally improve performance, this may allow to better accommodate for incorrect model specifications \citep[][\S 5.4.2]{rasmussen-williams-book}. 
Motivated by this observation, we explore this direction by extending the variational formulation to optimize a 
\loocv error. We also note that none of these previous \loocv approaches can actually handle large datasets.

\section{Gaussian Process Models \label{sec:gp-models}}
We are interested in supervised learning problems where we are given a training dataset of 
input-output pairs $\calD = \{ \x_n, \y_n\}_{n=1}^n$, with $\x_n$ being a $D$-dimensional
input vector and $\y_n$ being a $P$-dimensional output. 
Our  main goal is to learn a probabilistic mapping from inputs to outputs so that for a 
new input $\xstar$, we can estimate the probability of its associated label $p(\ystar | \xstar)$.

To this end, we follow a Gaussian process  modeling 
approach \citep[\gptext;][]{rasmussen-williams-book}, where 
latent functions are assumed to be distributed according to a \gptext, and 
observations are modeled via a suitable conditional likelihood given the latent functions.
A function $f_j$ is said to be distributed according to a Gaussian process with mean function 
$\meanfunc_j(\x)$ and covariance function $\kernel_j(\x, \x'; \hyperparam_j)$,
which we denote with $f_j \sim \gp(\meanfunc_j(\x), \kernel_j(\x, \x'; \hyperparam_j) )$,
if any subset of function values follow a Gaussian distribution. 
    
Since we are dealing with multiple outputs $\{\yn\}$, we follow the standard practice of considering
$Q$ underlying latent functions $\{f_j\}_{j=1}^{Q}$ which are drawn independently 
from zero-mean \gptext priors along with \iid conditional likelihood models. 
Such a modeling approach encompasses a large 
class of machine learning problems, including \gptext-regression, 
\gptext-classification and modeling of count data \citep{rasmussen-williams-book}. 

Consequently, when realized at the observed data, our probabilistic model is given by:
\begin{align}
\label{eq:prior}
	p(\f | \X, \hyperparam ) &= \prod_{j=1}^Q p(\fj | \X, \hyperparam_j) = \prod_{j=1}^Q \Normal(\fj; \vec{0}, \K_j) \text{,} \\
\label{eq:likelihood}
p(\y | \f ) &= \prod_{n=1}^N p(\y_n | \fn) \text{,}	
\end{align}
where $\f$ is the $(\n \times \q)$-dimensional vector of 
all latent function values; 
$\fj = \{f_j(\x_n)\}_{n=1}^N$ denotes the 
latent function values corresponding to the  $j\mth$ \gptext; 
$\K_j$ is the covariance matrix obtained by evaluating the
 covariance function $\kernel_j(\cdot,\cdot; \hyperparam)$ at all pairs of training inputs; 
  $\y$ is the vector of all $(\n \times \p)$ output observations,   
with $\y_n$ being $n\mth$ output observation, 
and $\fn = \{f_j(\x_n)\}_{j=1}^Q$ being the corresponding vector of latent function values. 
We refer to the covariance parameters $\hyperparam$ 
 as the hyperparameters.

One commonly used covariance function is the so-called squared exponential (or \rbf kernel):
\begin{equation}
	\label{eq:rbf-kernel}
	\kernel(\x, \x'; \hyperparam) = \sigma^2 \exp \left(- \sum_{i=1}^\d \frac{ (x_i - x'_i)^2 }{\ell_i^2}  \right) \textbf{,}
\end{equation}
where $\hyperparam = \{\sigma^2, \ell_1^2, \ldots, \ell_\d^2\}$ and $\d$ is the input 
dimensionality. When all the lengthscales, $\ell_i$, are constrained to be the same we refer to the above 
kernel as \emph{isotropic}; otherwise, we refer to it as the squared exponential covariance with 
automatic relevance determination (\ard).
\subsection{Inference Problems}
The main inference problems in \gptext models are (i) estimation of the posterior over the latent 
functions $p(\f | \X, \y, \hyperparam)$ and subsequent estimation of the predictive distribution 
$p(\ystar | \xstar, \X, \y, \hyper)$; and (ii) inference over the hyperparameters $\hyperparam$. 

For general likelihood models, both inference problems are analytically intractable, as they 
require the computation of non-trivial high-dimensional integrals. Furthermore, even for the 
simplest case when the conditional likelihood is Gaussian, evaluating the posterior and 
hyperparameter estimation is computationally prohibitive, as they entail time 
and memory requirements of $\bigO(\n^3)$ and $\bigO(\n^2)$ respectively.  

In addition to dealing with general likelihood models and scalability issues, the performance 
of \gptext models (and kernel machines in general), is highly dependent on the covariance function 
(or kernel) used. Most previous work on \gptext methods and \svm{s} in the machine learning literature,
 use the squared exponential kernel in Equation \eqref{eq:rbf-kernel}. 
 Furthermore, especially in high dimensions, this kernel is constrained to its isotropic version.  
As pointed out by \citet{bengio2005curse}, such an approach is severely limited by the 
curse of dimensionality when learning complex functions.  

Finally,  the estimation of 
hyper-parameters based solely on marginal likelihood optimization can be very sensitive to 
model misspecification, with predictive approaches such as those 
in \citet{sundararajan2001predictive} being seemingly more appealing
for this task, especially in problems such as classification, where we are  ultimately
interested in having lower error rates and better calibrated predictive probabilities.   

In the following sections, we show how to deal with the above issues, namely 
non-Gaussian likelihood models and scalability; more flexible kernels; and 
better objective functions for hyperparameter learning, with the endmost goal of improving performance significantly over  current
\gptext  approaches.

\section{Automated Variational Inference \label{sec:inference}}
In this section we detail our method for scalable and statistically efficient inference in 
Gaussian process models with general likelihoods as specified by Equations \eqref{eq:prior} and 
\eqref{eq:likelihood}. We adopt the variational framework 
of \citet{dezfouli-bonilla-nips-2015}, which we prefer over the \mcmc method of 
\citet{Hensman15b} as we avoid the sampling overhead which is especially significant in very large datasets. 
%
\subsection{Augmented Model}
In order to have a scalable inference framework, we augment our prior with $\m$ inducing 
variables $\{  \uj \}$ per latent process and corresponding inducing inputs $\{ \Zj\}$ such that,
\begin{align}
\label{eq:prior-u}
p(\u) & 
= \prod_{j=1}^\q \Normal(\uj; \vec{0}, \Kzz) \text{,}  \\
p(\f | \u) &= \prod_{j=1}^\q \Normal(\fj; \priormean, \priorcov) \text{, where }\\
\priormean &= \Kxz \Kzzinv \uj  \text{, and }\\
\label{eq:Ktilde-A}
\priorcov &= \Kxx - \Aj \Kzx \text{ with }
\Aj  = \Kxz \Kzzinv  \text{,}
\end{align}
where 
$\Gram{\vec{u}}{\vec{v}}$ is the covariance matrix obtained by evaluating 
the covariance function $\kernel_j(\cdot, \cdot; \hyperparam)$ at all 
pairwise columns of matrices $\mat{U}$ and $\mat{V}$. We note that 
this prior is equivalent to that defined in Equation \eqref{eq:prior}, which 
is obtained by integrating out $\vec{u}$ from the joint $p(\f, \u)$. However, 
as originally proposed by \citet{titsias2009variational}, having an explicit 
representation of $\u$ will allow us to derive a scalable variational framework 
without additional assumptions on the train or test conditional distributions 
\citep{quinonero2005unifying}.

We now define our approximate joint posterior distribution as:
\begin{align}
\label{eq:joint-posterior}
\qjoint & \defeq p(\f | \u) \qu \text{, with} \\	
\label{eq:var-posterior}
\qu &=  \sum_{k=1}^{\k} \pi_k \prod_{j=1}^\q  \Normal(\uj; \postmean{kj}, \postcov{kj}) \text{,}
\end{align}
where $p(\f | \u)$ is the conditional prior given in Equation \eqref{eq:prior-u} as variational parameters.
$\qu$ is our  variational posterior with $\llambda = \{ \pi_k, \postmean{kj}, \postcov{kj} \}$. 
Note that we assume a variational posterior in the form of a mixture for added flexibility compared to using a single distribution. 
\subsection{Evidence Lower Bound}
Such a definition of the variational posterior allows for simplification of the 
log-evidence lower bound ($\elbo$) such that no $\bigO(\n^3)$ operations are required, 
\begin{align}
\label{eq:elbo}
\nonumber
\elbo(\varparam, \hyperparam) &= - \kl{\qu}{p(\u)}  + \\
& \sum_{n=1}^\n \sum_{k=1}^\k \pi_k \expectation{\qknf}{ \log \pcond{\yn}{\fn} } \text{,}
\end{align}
where $\kl{q}{p}$ denotes the KL divergence between distributions $q$ and $p$; 
 $\expectation{p(X)}{g(X)}$ denotes the expectation of $g(X)$ over distribution $p(X)$;
  and $\qknf$ is a $\q$-dimensional diagonal Gaussian  with means and variances given by
\begin{align}
\label{mean-cov-postqf}
 \margmean{j} &= \ajn^T \postmean{kj}  \text{,} \qquad \\
 \margvar{j} &=  \matentry{ \priorcov }{n}{n} + \ajn^T \postcov{kj} \ajn \text{,}
\end{align}
where  $\ajn \defeq  \matcol{\Aj}{n}$   denotes the $\m$-dimensional vector corresponding to the $n\mth$ column of matrix
$\Aj$; $\priorcov$  and $\Aj$ are given in Equation \eqref{eq:Ktilde-A}; and,
$\matentry{ \priorcov }{n}{n}$ denotes the $(n,n)\mth$ entry of matrix $\priorcov$. 

In order to compute the log-evidence lower bound in  Equation \eqref{eq:elbo} and its gradients,  we 
use Jensen's inequality to bound the KL term  \citep[as in][]{dezfouli-bonilla-nips-2015} 
and estimate the expected likelihood 
term using   Monte Carlo (\mc).
%

\subsection{The Reparameterization Trick}
One of the key properties of the log-evidence lower bound in Equation \eqref{eq:elbo} is its 
decomposition as a sum of expected likelihood terms on the individual observations. This allows 
for the applicability of stochastic optimization methods and, therefore, scalability to very 
large datasets. 

Due to the expectation term in the expression of the $\elbo$, gradients of the $\elbo$ need to be estimated. 
\citet{dezfouli-bonilla-nips-2015} follow a similar approach to that 
in general black-box variational methods \citep{ranganath2014black}, and use the 
property $\grad_{\varparam} \expectation{q(\f | \varparam)}{\log p (\y | \f)} = 
\expectation{q(\varparam)}{\grad_{\varparam} \log q (\f | \varparam)  \log p (\y | \f)}$ and 
Monte Carlo (\mc) sampling to produce unbiased 
estimates of these gradients. While such an approach is truly black-box in that it does 
not require detailed knowledge of the conditional likelihood and its gradients, the estimated 
gradients may have significantly large variance, which could hinder the optimization process. In fact, 
\citet{dezfouli-bonilla-nips-2015} and 
\citet{bonilla2016lgpm} use variance-reduction techniques \citep[][\S8.2]{ross2006simulation}
  to ameliorate this effect. Nevertheless,  the experiments 
  in \citet{bonilla2016lgpm} on complex likelihood functions such as those in 
  Gaussian process regression networks \citep[\gprn{s};][]{wilson-et-al-icml-12} indicate that 
  such techniques may be insufficient to control the stability of the optimization process, and 
  a large number of \mc samples may be required. 
  
In contrast, if one is willing to relax the constraints of a truly black-box approach by having access to the 
implementation of the conditional likelihood and its gradients, then a simple trick proposed 
by \citet{kingma2013auto} can significantly reduce the variance in the gradients and make 
the stochastic optimization of $\elbo$ much more practical. This has come to be known as 
the \emph{reparameterization trick}.
 
We now present explicit expressions of our estimated $\elbo$, focusing on an individual 
 expected likelihood term in Equation \eqref{eq:elbo}.
 Thus, the individual expectations can 
 be estimated as:
\begin{align}
\epsilon_{knj} & \sim \Normal(0, 1) \text{,}\\
\label{eq:samples}
f^{(k,i)}_{nj} & = \margmean{j} + \margstd{j} \epsilon_{knj} \text{,} \quad j=1,  \ldots, \q \text{,}\\
\elltermhatn & = \frac{1}{\s}  \sum_{k=1}^\k \pi_k \sum_{i=1}^{\s} \log p(\yn | \fn^{(k,i)}) \text{,}
\end{align}
where $\s$ is the number of \mc samples and 
$f^{(k,i)}_{nj}$ is an element in the vector of latent functions $\fn^{(k,i)}$.
\subsection{Mini-Batch Optimization}
We can now obtain unbiased estimates of the gradients of $\elbo$ to use in mini-batch stochastic 
optimization. In particular, for a mini-batch $\batchset$ of size $\batchsize$ we have that 
the estimated gradient is given by:
\begin{align*}
\grad_{\param} \elbohat = -  \grad_{\param}  \kl{\qu}{p(\u)}  + 
	 \frac{\n}{\batchsize} \sum_{n \in \batchset} \grad_{\param} \elltermhatn \text{,}
\end{align*} 
where $\param \in \{\varparam, \hyperparam\}$;  $\hyperparam$ now includes not 
only the covariance parameters, but also the inducing inputs across all latent 
processes $\Zj$.
The corresponding
 gradients are obtained through automatic differentiation using TensorFlow.
\subsection{Full Approximate Posterior}
An important characteristic of our approach is that, unlike most previous work 
that uses stochastic optimization in variational inference along with the 
reparameterization trick  \citep[see e.g.][]{kingma2013auto,Dai15}, we do not 
require a fully factorized approximate posterior to estimate our gradients using only  
simple scalar transformations of uncorrelated Gaussian samples, as given in Equation \eqref{eq:samples}. 
Indeed,  the posterior  over the latent functions is fully correlated and is 
given in the Appendix.
%
This result is a property inherited from the original framework of 
 \citet{nguyen-bonilla-nips-2014} and \citet{dezfouli-bonilla-nips-2015}, who showed 
that, even when having a full approximate posterior, only expectations over univariate Gaussians are required in order to estimate 
the expected log likelihood term in  $\elbo$. 
%
%
%


\section{Flexible Kernels}
We now  turn our attention to increase the flexibility of \gptext models through the investigation of kernels beyond the isotropic \rbf kernel. 
This is a complementary direction for exploring the capabilities of \gptext models, but it benefits from the results of the previous 
section in terms of large-scale inference and computation of gradients via (automatic) back-propagation. 

For the \rbf kernel, we limit our experiments to the automatic relevance determination (\ard)
setting in Equation \eqref{eq:rbf-kernel}, including problems of large input dimensionality such as 
\mnist and \cifar. 
Furthermore, given the interesting results reported by \citet{chonips2009}, we also investigate their 
 \arccosine kernel, which was proposed as a kernel that mimics the computation of neural networks. 
 We give the mathematical details of the \arccosine kernel in the Appendix.

\section{Leave-One-Out  Learning \label{sec:loogp}}
 
 Here we focus on the average leave-one-out log predictive  probability for hyper-parameter learning,
  as an alternative to optimization of the marginal likelihood. 
  This can be particularly useful  in problems such as classification where one is mainly interested in 
  having lower error rates and better calibrated predictive probabilities. 
  
This objective function is obtained  by leaving one  datapoint out of the training set 
 and computing its  log predictive probability when training on the other points;
 the resulting values are then averaged across all the datapoints. Interestingly, in our 
 \gptext model this can be computed without the need for training $\n$ different models, as 
 the resulting expression is given by: 
 \begin{equation}
 \label{eq:loo}
\looobj(\hyperparam) \approx - \frac{1}{N} \sum_{n=1}^{N} \log  \int  \frac{q(\fn |  \calD, \varparam, \hyperparam)}{p(\yn | \fn)} \ d \fn \text{,}
\end{equation}
where the approximation stems from using the variational marginal posterior $q(\fn | \calD, \hyperparam)$ 
instead of the true marginal posterior  $p(\fn | \calD, \hyperparam)$ for datapoint $n$.
We also note that we have made explicit the dependency of the posterior on all the data. 
The derivation of this expression is given in the Appendix. Since $\looobj(\hyperparam)$ contains an expectation with respect to the marginal posterior we can also estimate it via \mc sampling.

Equation \eqref{eq:loo} immediately suggests an alternating optimization 
scheme where we estimate the approximate posterior $q(\fn | \calD, \hyperparam)$ through 
optimization of $\elbo$, as described in \S \ref{sec:inference}, and then 
learn the hyperparameters via optimization of $\looobj$.  
Algorithm \ref{alg:hyper} illustrates such an alternating scheme for hyperparameter learning
 in our model using the leave-one-out objective and 
vanilla stochastic gradient descent (\sgd), where 
$\elbohat$ and  $\looobjhat$ denote estimates of the corresponding 
objectives when using mini-batches.
%
\begin{algorithm}[t]
	\caption{$\looobj$-based hyperparameter learning. \label{alg:hyper}}
\begin{algorithmic}
\Repeat 
    \Repeat
        \State $(\hyper, \varparam) \gets (\hyper, \varparam) + \alpha \grad_{\hyper, \varparam} \elbohat(\hyperparam, \varparam)$
    \Until{satisfied}
    \Repeat 
        \State $\hyper \gets \hyper + \alpha \grad_{\hyper}  \looobjhat(\hyper)$
    \Until{satisfied}
\Until{satisfied}
\end{algorithmic}
\end{algorithm}


\section{Experiments}
We evaluate  \autogp across 
various datasets with number of observations ranging from $12,000$ to $8.1$M and 
input dimensionality between $31$ and $3072$. The aim of this section is (i) to demonstrate the effectiveness of the re-parametrization trick in reducing the number of samples needed to estimate the expected log likelihood; 
(ii) to evaluate non-isotropic kernels across a large number of dimensions; (iii) to assess the performance obtained by \loocv hyperparameter learning; and (iv) to analyze the performance of \autogp as a function of time.
Details of the experimental set-up can be found in the supplementary material.
%
%
\subsection{Statistical Efficiency}
For our first experiment we look at the behavior of our model as we vary the number of samples used to estimate the expected log likelihood. We trained 
our model on the \sarcos dataset \citep{vijayakumar2000locally}, an inverse dynamics problem for a seven degrees-of-freedom anthropomorphic robot arm. 

We used the Gaussian process regression network (\gprn) model of \citet{wilson-et-al-icml-12} as our likelihood function. 
\citet{bonilla2016lgpm} found that \gprn{}s require a large number of samples ($10,000$) to yield stable optimization of the $\elbo$ in their method. This is most likely due to the increased variance induced by multiplying latent samples together. The high variance of this likelihood model makes it an ideal candidate for us to evaluate how the performance of \autogp varies with the number of samples.

\begin{figure}[t]
		\centering
        \includegraphics[scale=1]{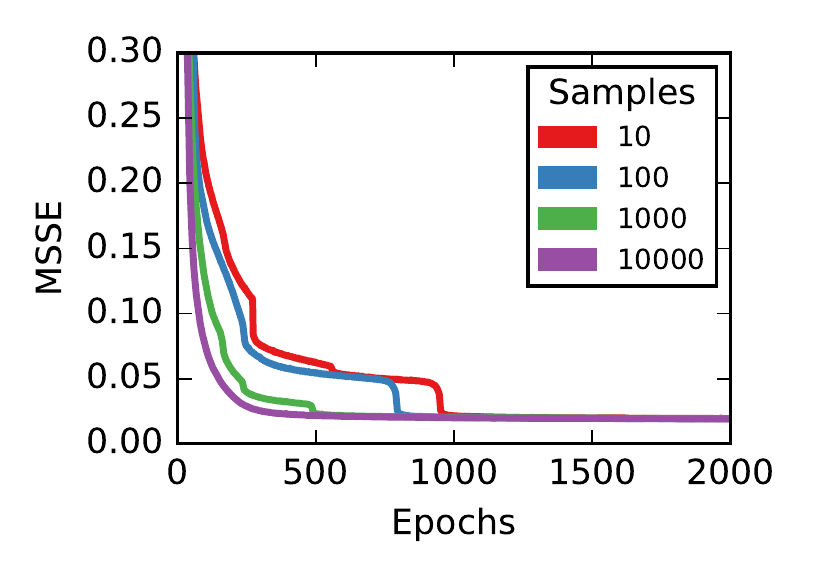}
        \caption{%
        The mean standardized square error (\msse) of \autogp for different number of \mc samples  on the \sarcos dataset. The \msse was averaged across all $7$ joints and $50$ inducing inputs were used to train the model. An epoch represents a single pass over the entire training set.
        }
        \label{fig:sarcos}%
\end{figure}

As shown in Figure~\ref{fig:sarcos}, increasing the number of samples decreased the number of epochs it took for our model to converge; however, our model always converged to the same optimal value. This is in stark contrast to 
\savigp \citep{bonilla2016lgpm} which was able to converge to the optimal value of $0.0195$ with $10,000$ samples, but converged to sub-optimal values ($>0.05$) as the number of samples was lowered.  
This shows that our \mc estimator is significantly more stable than \savigp{}'s black-box estimator, without requiring any variance reduction methods. In practice, reducing the number of samples leads to an improved runtime. A gradient update using $10,000$ samples takes around $0.17$ seconds, whereas an update using $10$ samples only takes around $0.03$ seconds, which makes up for the extra epochs needed to converge. 

\subsection{Kernel Performance}\label{sec:kernel}
In this section we compare the performance of \autogp across two different kernels on a high-dimensional dataset. We trained our model on the \rectangle dataset \citep{larochelle07} which is a binary classification task that was created to compare shallow models (e.g.~\svm{}s), and deep learning architectures. The task involves recognizing whether a rectangle has a larger width or height. Randomized images were overlayed behind and over the rectangles, 
which makes this task particularly challenging.
%

We compare the multilayer arc-cosine kernel (\arccosine) described by \citet{chonips2009} with an \rbf kernel with automatic relevance determination (\rbfard). \arccosine was devised to mimic the architecture of deep neural networks through successive kernel compositions, and has shown good results when combined with non-Bayesian kernel machines \citep{chonips2009}. Unlike \arccosine{}, \rbfard is commonly used with Gaussian processes \citep{hensmangaussian}. However, \rbfard has not been applied to large-scale datasets with a large input dimensionality due to limitations in scalability. Given that our implementation uses automatic differentiation, and is thus able to efficiently compute gradients, we can use \rbfard with no noticeable performance overhead.

We trained our model using $10$, $200$, and $1000$ inducing points across both kernels. We experimented with various different depths and degrees for \arccosine and found that $3$ layers and a degree of $1$ worked best with \autogp. As such, we ran all \arccosine experiments with these settings.

For reference, \citet{chonips2009} report an error rate of $22.36\%$ using \svm{}s with \arccosine. \citet{larochelle07} report error rates of $24.04\%$ on \svm{}s with a simple \rbf kernel and $22.50\%$ with a deep belief network with $3$ hidden layers (\dbnthree). 

\begin{figure}[t]
	\centering
		\includegraphics[scale=1]{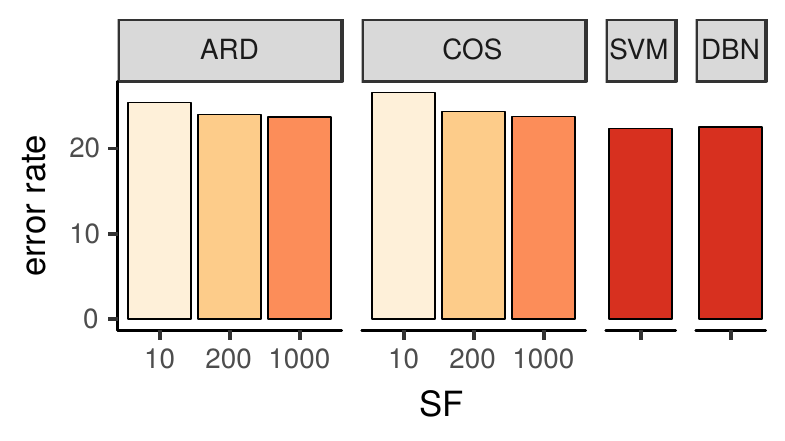}
        \caption{%
        Error rates for binary classification using a logistic likelihood  on  \rectangle. ARD and COS refer to  \autogp with a \rbfard kernel, and a $3$-layer \arccosine kernel of degree $1$ respectively. 
        \dbnthree refers to a $3$-layer deep belief network and \svm denotes a support vector machine with a \arccosine kernel of depth $6$ and degree $0$.
        }
        \label{fig:kernel}%
\end{figure}

	Figure~\ref{fig:kernel} shows that \autogp is competitive with previous approaches, except when using  $10$ inducing points. Our results also show that \rbfard performs comparatively (and sometimes better) than \arccosine.  
	This is likely because \arccosine is better suited to deep architectures, such as the multilayer kernel machine of  \citet{chonips2009}.

\subsection{LOO-CV Hyperparameter Learning }
In this section we analyze our approach when learning 
hyperparameters via optimization of the leave-one-out objective  ($\looobjhat$), 
as described in \S \ref{sec:loogp}, 
and compare it with the standard variational method that learns hyperparameters via sole optimization 
of the variational objective ($\elbohat$). To this end, we use the \mnist dataset 
 using $10$, $200$, and $1000$ inducing points and  the \rbfard  kernel across all settings.

\begin{figure}[t]
        \centering
        \includegraphics[scale=1]{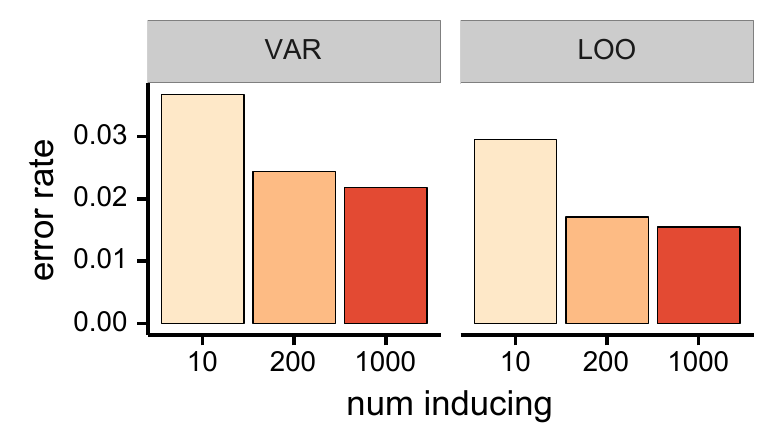}
        \caption{%
        Error rates of \autogp for multiclass classification using a softmax likelihood  on  \mnist. 
        VAR refers to learning using only the 
        variational objective $\elbohat$, while LOO refers to learning 
        hyperparameters  using the leave-one-out objective $\looobjhat$.
        }
        \label{fig:mnist_er}%
\end{figure}

Our results indicate that optimizing the leave-one-out objective leads to a significant performance gain.
While Figure \ref{fig:mnist_er} shows the corresponding error rates, we refer the reader to the supplementary 
material for similar results on negative log probabilities (\nlp{s}).
In summary, at $1000$ inducing points our \loocv approach attains an error rate of $1.55\%$ and a mean \nlp of $0.061$; 
At $200$ inducing points it obtains an error rate of $1.71\%$ and a mean \nlp of $0.063$. 
This represents a $40\%$--$45\%$ decrease in mean \nlp and error rate from the equivalent model 
that only optimized the variational objective. 
Furthermore, our results  outperform the work of \citet{Hensman15b},  \citet{gal-et-al-nips-2014}, 
and \citet{bonilla2016lgpm}  who reported  error rates of $1.96\%$,  $5.95\%$ and  $2.74\%$  respectively, that are the best results reported on this data set in the literature of GPs. 

We have shown that our approach reduces the error rate on the test set down to $1.55\%$ without artificially extending the \mnist dataset through transformations, or doing any kind of pre-processing of the data. These results, while not on par with state-of-the-art vision results, show that Gaussian process models can perform competitively with non-Bayesian methods, while solving the harder problem of full posterior estimation. For example, standard convolutional neural networks achieve an error rate of $1.19\%$ without any pre-processing \citep{img-results}. We will show in the following section  how we can further improve our results by artificially extending the training data on \mnist through distortions and affine transformations.

\subsection{AutoGP All Together}
We finish by evaluating the performance of \autogp using all its main components  simultaneously, namely 
statistical efficiency, flexible kernels and \loocv hyperparameter learning, using three  datasets: \rectangle, \cifar, and \mnisteight.
 Our goal is to gain further insights as to how our model compares against other works in the machine learning community. 
 As before, we optimized our model with $200$ inducing points and used the \rbfard kernel across all experiments.

We begin by looking at the \rectangle dataset which we have already investigated in Section~\ref{sec:kernel}. 
Optimizing $\looobjhat$ slightly improves performance in this case, with a reduction in error rate from $24.06\%$ to $24.04\%$ and a reduction in mean \nlp from $0.495$ to $0.485$, when using 200 inducing points. The error rate can be further reduced to $23.6\%$ when using 1,000 inducing points (with corresponding \nlp of 0.478).  As a reference, performance of recent deep architectures is: deep trans-layer autoencoder networks \citep[$\er=13.01\%$,][]{zhu15}, PCANet \citep[$\er=13.94\%$,][]{chang14}, and two-layer stacked contractive auto-encoders \citep[$\er=21.54\%$,][]{rifai2011}.  \citet{bruna13} have reported results using invariant scattering convolutional networks with an error rate of $8.01\%$.

\begin{figure}[t]
		\centering
        \includegraphics[scale=0.8]{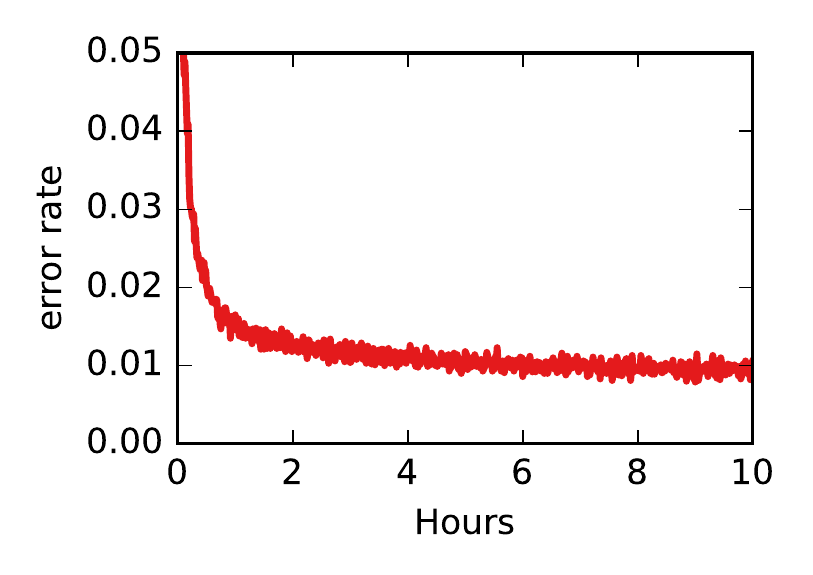}
        \caption{%
        Error rate of \autogp over time for multiclass classification using a softmax likelihood  on  \mnisteight. Training was done  with $200$ inducing points and $\looobj$ 
        for hyperparameter learning.
    }
        \label{fig:mnist_time}%
\end{figure}

We now turn our attention to the \mnisteight \citep{loosli-canu-bottou-2006} dataset, which artificially extends the \mnist dataset to $8.1$ million training points by pseudo-randomly transforming existing \mnist images. To the best of our knowledge, no other \gptext models have been shown to scale to such a large dataset. We were able to achieve an error rate of $0.89\%$ and a mean \nlp of $0.033$, breaking the $1\%$ error-rate barrier in \gptext models. 
This shows that our model is able to perform on par with deep architectures with some pre-processing of the \mnist dataset. We also note that we were able to rapidly converge given the dataset size. As shown in Figure~\ref{fig:mnist_time} \autogp achieves near optimal results within the first hour, and converges to the optimal value within 10 hours.

Finally, we look at \cifar{}, a multiclass image classification task and we find that results are less conclusive. We achieve an error rate of $44.95\%$ and a mean \nlp of $1.33$, 
which is well below the state of the art, with some models achieving less than $10\%$ error rate. 
However, we note that we still compare favorably to \svm{}s which have been shown to achieve error rates of $57.7\%$ \citep{Le13}. 
We also perform on par with older deep architectures \citep{rifai2011}. 
We believe these results provide a strong motivation for investigating application-specific kernels 
\citep[such as those studied by][]{mairal2014convolutional} in \gptext models.

\begin{table}[h]
\caption{Test error rate and mean \nlp of \autogp trained with $200$ inducing points on various datasets.  The $\looobjhat$ objective was used for hyperparameter learning} \label{model-table}
\begin{center}
\begin{tabular}{llll}
	\toprule
{Dataset}  & Error rate & mean \nlp \\
\midrule
\rectangle & 24.06\% & 0.485 \\
\mnisteight & $0.89\%$ & $0.033$ \\
\cifar & $44.95\%$ & $1.333$ \\
\bottomrule
\end{tabular}
\end{center}
\end{table}

\section{CONCLUSIONS \& DISCUSSION}
We have developed \autogp, an inference framework for \gptext models 
that pushes the performance  of current \gptext methods 
by exploring three complementary directions: (i) scalable and 
statistically efficient variational inference; (ii) flexible kernels; and 
(iii) objective functions for hyperparameter learning alternative to the marginal likelihood. 

%
We have 
shown that our framework outperforms all previous \gptext approaches
on \mnist; 
achieves comparable performance to previous kernel approaches on the \rectangle dataset; 
and can break the $1\%$ error barrier using \mnisteight. Overall, this represents 
a significant breakthrough in Gaussian process methods and kernel machines. 
While our results on \cifar are well below the state-of-the-art achieved with deep learning,
 we believe we can further reduce the gap between kernel machines and deep architectures 
 by using application-specific kernels such as those recently proposed by 
 \citet{mairal2014convolutional}.


\appendix
\section{Details of Full Posterior Distribution}
The full posterior distribution over the latent functions is given by:
\begin{align} 
\label{eq:qf}
\qf &=   \sum_{k=1}^\k \pi_k \prod_{j=1}^\q \Normal(\fj; \qfmean{kj}, \qfcov{kj}) \text{, where}\\
\label{eq:meanqf}
\qfmean{kj} & = \Aj \postmean{kj} \text{, and } \\
\label{eq:covqf}
\qfcov{kj} &= \priorcov + \Aj \postcov{kj} \Aj^T \text{.}
\end{align}

\section{Details of The Arc-Cosine Kernel}
\citet{chonips2009} define an \arccosine kernel of degree $\degree$ and depth $\depth$ using the 
following recursion:

\begin{align}
\arckernel{\depth + 1}{\degree}{\x}{\x'} &= 
\frac{1}{\pi} \left( \arckernel{\depth}{\degree}{\x}{\x} 
\arckernel{\depth}{\degree}{\x'}{\x'}
\right)^{\degree/2} J_{\degree}(\arcangle^{(\depth)}_{\degree}) \\
\arckernel{1}{\degree}{\x}{\x'} &= \frac{1}{\pi} \norm{\x}^{\degree} \norm{\x'}^{\degree} J_{\degree}
(\arcangle)
\end{align}
where 
\begin{align}
\arcangle^{(\depth)}_{\degree} 
&= \cos^{-1} \left(  \arckernel{\depth}{\degree}{\x}{\x'} 
\left( \arckernel{\depth}{\degree}{\x}{\x} 
\arckernel{\depth}{\degree}{\x'}{\x'} \right)^{-1/2} \right)  \\
J_{\degree} (\arcangle) & = (-1)^{\degree} (\sin \arcangle)^{2 \degree + 1} 
\left(\frac{1}{\sin \arcangle} \deriv{}{\arcangle} \right)^{\degree} \left(\frac{\pi - \arcangle}{\sin \arcangle} \right)\\
\arcangle &= \cos^{-1} \left( \frac{\x \cdot \x'}{\norm{\x} \norm{\x'}} \right) \text{.}
\end{align}
\section{Derivation of Leave-One-Out Objective}
In this section we derive an expression for the leave-one-out objective and show that this does 
not require training of $\n$ models. A similar derivation can be found in \citet{vehtari2016bayesian}. 
Let $\Dnotn = \{ \Xnotn, \ynotn\}$ be the dataset resulting from 
removing observation $n$.  Then our leave-one-out objective is given by:
\begin{align}
\label{eq:loo-app}
\looobj(\hyperparam) =  \frac{1}{N} \sum_{n=1}^{N} \log p(\yn | \x_n, \Dnotn, \hyper) \text{.}
\end{align}
We now that the marginal posterior can be computed as:
\begin{align}
	p(\fn | \calD) = p(\fn | \Xnotn, \ynotn, \x_n, \yn)  = \frac{p(\yn | \fn) p(\fn | \x_n, \Dnotn)} {p(\yn | \x_n, \Dnotn, \hyper) }
\end{align}
and re-arranging terms 
\begin{align}
 \int p(\fn | \x_n, \Dnotn, \hyper) d\fn & = \int \frac{p(\fn | \calD, \hyper)  p(\yn | \x_n, \Dnotn, \hyper)}{ p(\yn | \fn)} d\fn \\
  p(\yn | \x_n, \Dnotn, \hyper) & = {1}/{\int \frac{p(\fn | \calD, \hyper)}{ p(\yn | \fn)} d\fn } \\
  \log p(\yn | \x_n, \Dnotn; \hyper) & = - \log \int \frac{p(\fn | \calD, \hyper)}{ p(\yn | \fn)} d\fn \text{,} 
\end{align}
and substituting this expression in Equation \eqref{eq:loo-app} we have
\begin{align}
\looobj(\hyperparam) &=   - \frac{1}{N} \sum_{n=1}^{N} \log  \int p(\fn | \calD, \hyper) \frac{1}{p(\yn | \fn)} \ d \fn \text{.}
\end{align}
We see that the objective only requires estimation of the marginal posterior $p(\fn | \calD, \hyper)$, which we can 
approximate using variational inference, hence:
\begin{align}
\looobj(\hyperparam) \approx   - \frac{1}{N} \sum_{n=1}^{N} \log  \int q(\fn | \calD, \hyper) \frac{1}{p(\yn | \fn)} \ d \fn \text{,}
\end{align}
where $q(\fn | \calD, \hyper)$ is our approximate variational posterior. 
\section{Additional Details of Experiments}
\subsection{Experimental Set-up}
The datasets used are described in Table \ref{tab:datasets}. 
\begin{table}[t]
	\begin{center}
		\caption{The datasets used in the experiments and the corresponding
			models used. $\n_{train}, \n_{test}, \d$ are the number of training points, test points and input dimensions respectively.
		}
		\label{tab:datasets}
		\begin{tabular}{r l l l  l}
			\toprule
			Dataset & \textbf{$\n_{train}$} & \textbf{$\n_{test}$} & \textbf{$\d$} & Model   \\
			\midrule
			\sarcos & $44,484$ & $4,449$ & $21$  & \gprn \\
			\rectangle & $12,000$ & $50,000$ & $784$  & Binary classification \\
			\mnist & $60,000$ & $10,000$ & $784$  & Multi-class classification \\
			\cifar & $50,000$ & $10,000$ & $3072$  & Multi-class classification \\
			\mnisteight & $8.1$M & $10,000$ & $784$  & Multi-class classification  \\
			\bottomrule
		\end{tabular}
	\end{center}
\end{table}
We trained our model stochastically using the \rmsprop optimizer provided by TensorFlow \citep{tensorflow2015-whitepaper} with a learning rate of $0.003$ and mini-batches of size $1000$. We initialized inducing point locations by using the k-means clustering algorithm, and initialized the posterior mean to a zero vector, and the posterior covariances to identity matrices. When jointly optimizing $\looobjhat$ and $\elbohat$, we alternated between optimizing each objective for $100$ epochs. Unless otherwise specified we used $100$ Monte-Carlo samples to estimate the expected log likelihood term.

All timed experiments were performed on a machine with an Intel(R) Core(TM) i5-4460 CPU, 24GB of DDR3 RAM, and a GeForce GTX1070 GPU with TensorFlow 0.10rc.
\subsection{Additional Results}
\begin{figure}[t]
	\centering
	\includegraphics[scale=1]{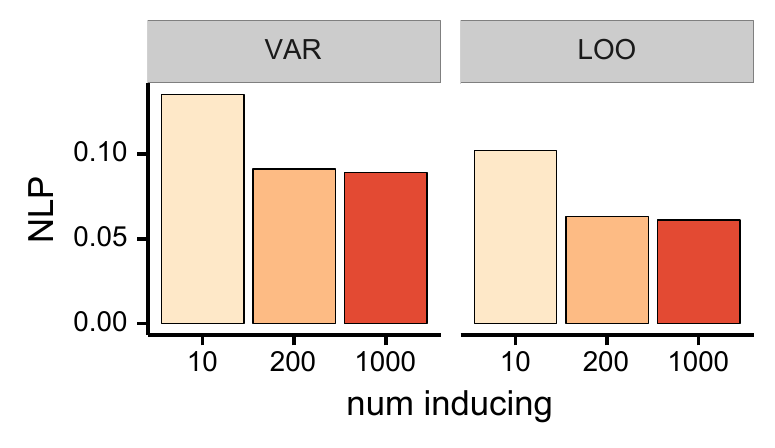}
	\caption{%
		\nlp for multiclass classification using a softmax likelihood model on the \mnist dataset.
		VAR shows the performance of \autogp where all parameters are learned using only the 
		variational objective $\elbohat$, while LOO represents the performance of \autogp when 
		hyperparameters are learned using the leave-one-out objective $\looobjhat$.
	}
	\label{fig:mnist_nlp}%
\end{figure}

Figure \ref{fig:mnist_nlp} shows the \nlp for our evaluation of the \loocv-based hyperparameter learning. As with the error rates described in the main text, 
the \nlp obtained with \loocv are significantly better than those obtained with a purely variational approach. 

\bibliographystyle{plainnat}
\bibliography{main-autogp}
\end{document}